\documentclass[letterpaper, 10 pt, conference]{ieeeconf}  

\IEEEoverridecommandlockouts                              

\usepackage{graphics} 
\usepackage{epsfig} 
\usepackage{mathptmx} 
\usepackage{times} 
\usepackage{amsmath} 
\usepackage{amssymb}  

\usepackage{amsmath,amssymb,amsfonts}
\usepackage{graphicx}
\usepackage{url}       
\usepackage{booktabs}
\usepackage{multirow}
\newcommand{\technique}{\textit{SCGT}}

\title{\LARGE \bf
SELF-CLUSTERING GRAPH TRANSFORMER APPROACH TO MODEL RESTING STATE
FUNCTIONAL BRAIN ACTIVITY
}

\author{Bishal Thapaliya $^{1,2}$, Esra Akbas$^{1}$,  Ram Sapkota $^{1,2}$  Bhaskar Ray $^{1,2}$ Vince Calhoun $^{2,3}$  Jingyu Liu $^{1,2}$%
\thanks{*This work was supported by NIH grant R01AG063153.}
\thanks{$^{1}$ Department of Computer Science, Georgia State University, Atlanta, USA}%
\thanks{$^{2}$ Tri-Institutional Center for Translational Research in Neuroimaging and Data Science}%
\thanks{$^{3}$ School of Electrical and Computer Engineering, Georgia Institute of Technology, Atlanta, USA}%
}

\begin{document}

\maketitle
\thispagestyle{empty}
\pagestyle{empty}

\begin{abstract} 

Resting-state functional magnetic resonance imaging (rs-fMRI) offers valuable insights into the human brain's functional organization and is a powerful tool for investigating the relationship between brain function and cognitive processes as it allows for the functional organization of the brain to be captured without relying on a specific task or stimuli. In this study, we introduce a novel attention mechanism for graphs with subnetworks, named Self Clustering Graph Transformer (\technique), designed to handle the issue of uniform node updates in graph transformers. By using static functional connectivity (FC) correlation features as input to the transformer model, \technique\ effectively captures the sub-network structure of the brain by performing cluster-specific updates to the nodes unlike uniform node updates like vanilla graph transformers, further allowing us to learn and interpret the subclusters. We validate our approach on the Adolescent Brain Cognitive Development (ABCD) dataset, comprising 7,957 participants, for the prediction of total cognitive score and gender classification. Our results demonstrate that \technique\ outperforms the vanilla graph transformer method, and other recent models, offering a promising tool for modeling brain functional connectivity and interpreting the underlying subnetwork structures.

\end{abstract}
\section{Introduction}
\label{sec:intro}

Resting-state functional magnetic resonance imaging (rs-fMRI) provides an unprecedented opportunity to delve into the human brain's functional organization \cite{Lee2012}. By measuring spontaneous fluctuations in the blood-oxygen-level-dependent (BOLD) signal when the brain is at rest, rs-fMRI reveals intrinsic functional connectivity patterns among different brain regions. Although traditional MRI studies in the literature focused on structural brain measures for various phenotypes (\cite{Ram_Sapkota2024, Suresh2023, Ray2023}), rapidly growing studies have emerged to investigate the prediction of intelligence based on brain functional features (\cite{Thapaliya_DSAM, Ferguson2017,He2020,Dubois2018}). Functional connectivity (FC) is defined as the degree of temporal correlation between regions of the brain computed using time series of BOLD signals, and rs-fMRI FC provides a comprehensive view of the brain’s intrinsic organization.

Graph neural networks (GNNs) have emerged as powerful tools for modeling complex relational data, such as brain networks (\cite{Smith2013, Thapaliya2024}). However, most existing GNNs assume uniform node updates across the entire graph, which poses challenges when dealing with the intricate subnetwork structures of the brain connectome. Recall that GNNs update each node’s embedding using information from the node’s neighbors. The information
exchange is mediated by trainable weight matrices. The same weights are uniformly applied to all nodes. However, such uniform node updates can make the GNN overlook rich local structures in the graph. Nodes within the same cluster exhibit higher similarity and stronger dependencies than nodes from different clusters. This case applies completely to Graph Transformers (GTs) as well. Current GNNs and GTs can miss these nuanced local patterns and community-specific behaviors by treating all nodes identically. Addressing this issue, BrainGNN \cite{Li2021} and BrainRGIN \cite{Thapaliya2025} introduced a clustering-based embedding technique within the graph convolutional layer, enabling nodes belonging to distinct clusters to learn embeddings in a customized manner. However, there remains a need for attention mechanisms, more powerful than graph convolutions, that can effectively handle subclustered graphs and allow for the interpretation of learned subclusters.

In this study, we propose a novel attention mechanism for graphs with subnetworks, named Self Clustering Graph Transformer (\technique), designed to address the issue of uniform node updates in GTs. By using static FC correlation features as input to the transformer model, our approach leverages a self-clustering graph attention block to capture the subnetwork structure of the brain. This allows us to learn and interpret the subclusters within the brain's functional networks. Unlike traditional transformer models, which apply uniform attention mechanisms across all nodes, \technique\ employs a specialized attention mechanism tailored for subclustered graphs.

We validate our approach using the Adolescent Brain Cognitive Development (ABCD) dataset, comprising 7,957 participants, for the prediction of total cognitive score and gender classification. Our results demonstrate that \technique\ outperforms the vanilla graph transformer method and is competitive with other baseline models, offering a promising tool for modeling brain functional connectivity and interpreting the underlying subnetwork structures.

\section{Materials and Methods}

\subsection{The ABCD Dataset}

The Adolescent Brain Cognitive Development (ABCD) study is a large ongoing longitudinal study following youths from age 9-10 into late adolescence to understand factors that influence individual brain development. Participants were recruited from 21 sites across the United States to represent various demographic variables. Data used in this study were from 7,957 children aged 9–10 at baseline, including rs-fMRI images, total cognitive composite scores measured by the NIH Toolbox Cognitive Battery, and gender information. Data were split into training (n = 4773), validation (n = 1592), and test (n = 1592) subsets. Age and site effects were regressed out from the cognitive scores to ensure that only the relevant intelligence features are captured by our graph model.

We conducted preprocessing on the raw rs-fMRI data utilizing a combination of  FSLv6.0 and SPM12 toolboxes, encompassing several key steps, namely: 1) rigid head motion correction; 2) distortion correction; 3) removal of dummy scans; 4) normalization to the Montreal Neurological Institute space; and 5) a 6mm Gaussian kernel smoothing. Subsequently, we employed a fully automated spatially constrained independent component analysis to extract 100 intrinsic components using the Neuromark\_fmri\_1.0 template \cite{Du2020} as the 100 nodes of graph models. The static functional connectivity (FC) was computed as the Pearson correlation coefficients between the time series of each pair of nodes, resulting in a 100 $\times$ 100 correlation matrix for each subject.

\subsection{Proposed Architecture}

The architecture of \technique\ is designed to effectively model the subnetwork structures within the brain by introducing a novel attention mechanism tailored for subclustered graphs. Our model uses static FC correlation features as input to the transformer model. The input graph is constructed with nodes representing brain regions of interest (ROIs) and edge weights representing the correlation coefficients between the ROIs. Positional encodings are incorporated using Laplacian eigenvectors to capture the structural information of the graph.

The core component of \technique\ is the self-clustering graph attention block, which enables nodes belonging to different clusters of the brain to learn embeddings in a customized manner, addressing the issue of uniform node updates in traditional GNNs and GTs. Unlike standard transformer models that apply the same attention mechanism across all nodes, our approach allows for specialized attention mechanisms for each subnetwork or cluster within the graph. The output node representations from the attention blocks are then aggregated using a concatenation readout function, which combines the features from all nodes to produce a graph-level representation. This representation is then passed through a fully connected layer for the final prediction or classification task.

  \begin{figure}[ht]
\includegraphics[width=\columnwidth]{ 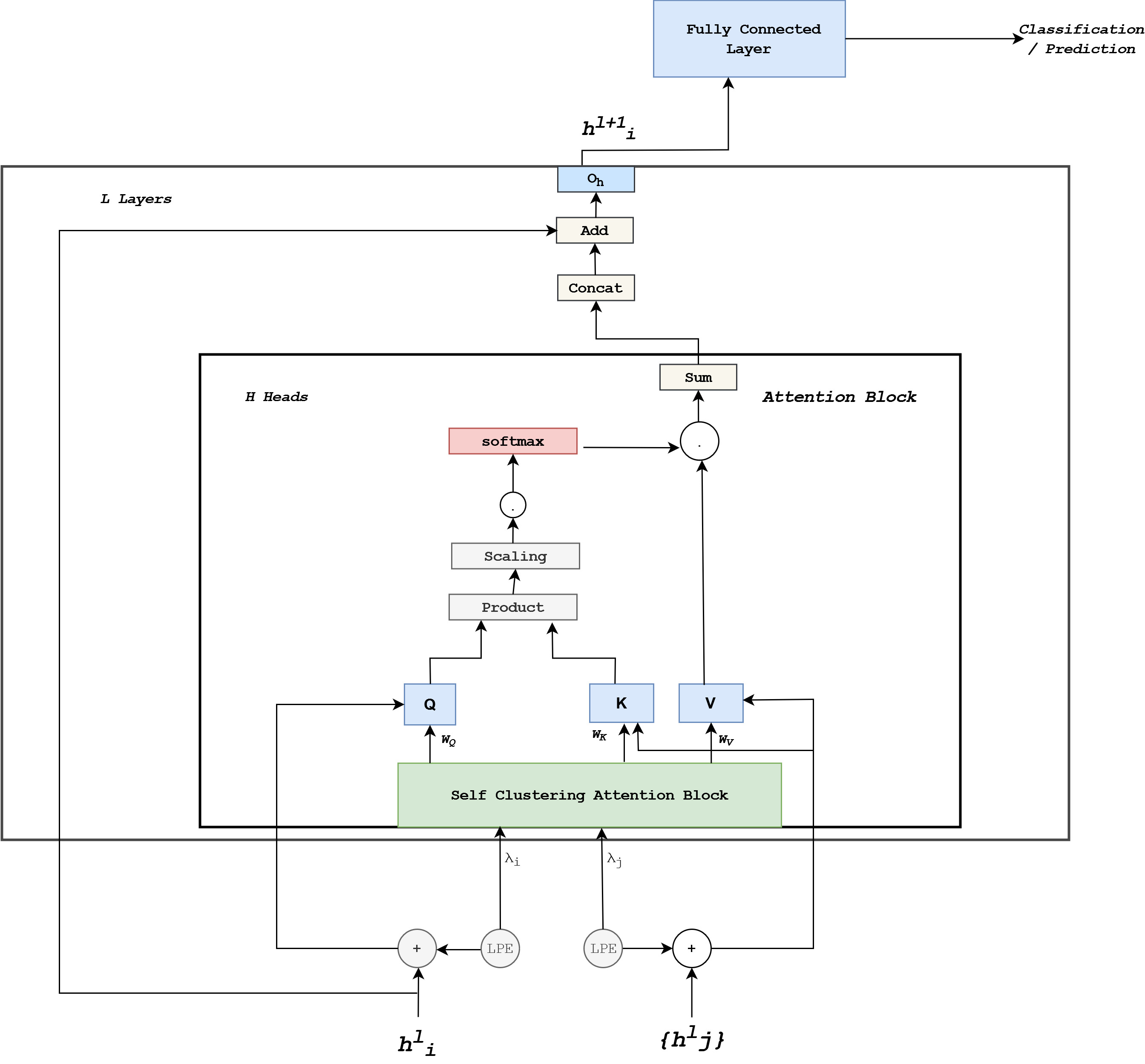}
\caption{Overall Architecture of \technique. The static functional connectivity features are used as input to the self-clustering graph attention blocks. Positional encodings are incorporated using Laplacian eigenvectors. The output node representations are concatenated to form the graph-level representation for prediction tasks.}
\label{SCGT_Architecture}
\end{figure}



\subsection{Functional connectivity graph creation}

The initial phase involves the processing of input from fMRI time series data to create a static FC matrix. Using ROIs time series, Pearson's correlation between ROIs is used to create a FNC matrix. The resulting FNC matrix is used to construct an undirected graph denoted by the tuple \begin{math} G = (V, E) \end{math}  with an adjacency matrix $A \in R^{N*N}$.
Each ROIs represents graph nodes indexed by the set \begin{math} V = 1,..., N \end{math}. The input node features,  \begin{math} h_i\end{math}, are formed using Pearson's correlation coefficients between node i and all other nodes. An edge set \begin{math} E\end{math} represents the functional connections between ROIs with each edge  \begin{math} e_{ij} \end{math} linking two nodes \begin{math} (i,j) \in E \end{math}. 

The Adjacency matrix $A_k \in R^{N \times N}$ is formed as in Equation \ref{eq:adj}, where $e_{i,j} $  is the thresholded element of FNC matrix to achieve either fully connected or sparse graph. 

\begin{equation}
    A_{i,j}^k = \left\{
    \begin{array}{ll}
        0, & i=j \\
        e_{i,j}, & \text{otherwise}
    \end{array} \right.
\label{eq:adj}
\end{equation}

\subsubsection{Laplacian Eigenvectors as Positional Encodings}

To include positional encodings (PE), we make use of Laplacian eigenvectors, a fundamental concept in spectral graph theory, to represent the underlying graph structure \cite{GraphTransformer}. Laplacian eigenvectors capture the global structure of the graph and provide meaningful position information for each node. The normalized graph Laplacian is computed as:

\[
L = I - D^{-\frac{1}{2}} A D^{-\frac{1}{2}}
\]

where $D$ is the degree matrix, and $A$ is the adjacency matrix. The eigenvectors $\lambda_i$ corresponding to the smallest non-trivial eigenvalues are used as positional encodings for node $i$.

\subsubsection{Self-Clustering Graph Attention Block}

The self-clustering graph attention block is a pivotal component of our architecture, specifically designed for brain analysis. Unlike conventional GT models that apply uniform processing to all nodes, this block incorporates specialized mechanisms to account for the brain's functional clustering. It leverages domain-specific knowledge about the brain's functional clusters, recognizing the distinct divisions among brain regions and their unique interactions.
\paragraph{Input Representation}

The input to the self-clustering graph attention block consists of node features and positional encodings:

\begin{align}
\hat{h}_i &= h_i + \lambda_i,
\end{align}

where $h_i$ is the node feature, and $\lambda_i$ is the positional encoding for node $i$.

\paragraph{Self-Clustering Attention}

The attention mechanism within this block is tailored to accommodate the brain's sub-clustered structure. The parameters for the key, query, and value matrices in layer $l$ are defined as functions of the positional encoding $\lambda_i$:

\begin{align}
\mathbf{w(k)}^{(l)}_i &= \theta^{(l)}_{k2} \cdot \text{ReLU}\left(\theta^{(l)}_{1} \cdot \lambda_i\right) + b_k^{(l)}, \\
\mathbf{w(q)}^{(l)}_i &= \theta^{(l)}_{q2} \cdot \text{ReLU}\left(\theta^{(l)}_{1} \cdot \lambda_i\right) + b_q^{(l)}, \\
\mathbf{w(v)}^{(l)}_i &= \theta^{(l)}_{v2} \cdot \text{ReLU}\left(\theta^{(l)}_{1} \cdot \lambda_i\right) + b_v^{(l)},
\end{align}

where $\theta^{(l)}_{1} \in \mathbb{R}^{N \times k_r}$ is a shared parameter matrix across all key, query, and value computations, facilitating the learning of clustering assignments for the nodes. Here, $k_r$ denotes the number of clustered communities.

Using these parameters, the key, query, and value matrices are computed as:

\begin{align}
K_{ROI} &= \mathbf{W(K)}^{(l)} \cdot \mathbf{\hat{H}}, \\
Q_{ROI} &= \mathbf{W(Q)}^{(l)} \cdot \mathbf{\hat{H}}, \\
V_{ROI} &= \mathbf{W(V)}^{(l)} \cdot \mathbf{\hat{H}},
\end{align}

where $\mathbf{\hat{H}}$ is the matrix of node features with positional encodings.

\paragraph{Layer Update Equations}

The update mechanism for node features in layer $l$ is governed by the following equation:

\begin{align}
\hat{h}_{i}^{(l+1)} &= \mathcal{O}_{h}^{(l)} \left[ \hat{h}_i^{(l)} + \bigg\Vert_{k=1}^{K} \left( \sum_{j \in \mathcal{N}_i} w_{ij}^{k,l} \cdot V^{(l)} \cdot \hat{h}_{j}^{(l)} \right) \right],
\end{align}

where:
\begin{itemize}
    \item $\mathcal{O}_{h}^{(l)}$ is an output transformation function for node features, typically a linear layer followed by a non-linear activation.
    \item $K$ denotes the number of attention heads or clusters.
    \item $\mathcal{N}_i$ is the set of neighboring nodes connected to node $i$.
    \item $w_{ij}^{k,l}$ are the attention weights for the $k$-th head at layer $l$, computed as:
    \begin{align}
    w_{ij}^{k,l} &= \text{softmax}_j\left( \frac{Q^{(l)}_{ik} \cdot K^{(l)}_{jk}}{\sqrt{d_k}} \right),
    \end{align}
    where $Q^{(l)}_{ik}$ and $K^{(l)}_{jk}$ are the elements of the query and key matrices for nodes $i$ and $j$ in head $k$ at layer $l$, respectively, and $d_k$ is the dimensionality of the key/query vectors.
\end{itemize}






\vspace{-1.5em}
\subsection{Readout}

After passing through the self-clustering graph attention blocks, the node representations are aggregated using a concatenation readout function. This function combines the features from all nodes into a single graph-level representation:

\begin{equation}
h_G = \text{Concat}(\hat{h}_1, \hat{h}_2, ..., \hat{h}_N)
\end{equation}

This graph-level representation $h_G$ is then passed through a fully connected layer for the final prediction or classification task.

\section{Experimental Setup and Results}

\subsection{Training the Model}

The neural network architecture depicted in Fig. \ref{SCGT_Architecture} was implemented using PyTorch \cite{PazkeTorch} and PyTorch Geometric \cite{TorchGeometric} for the specific graph neural network components. The number of nodes was 100 (corresponding to the Neuromark ICA components). We used 100 eigenvectors as positional encodings $\lambda_i$ for each node $i$. We used two layers of self-clustering graph attention blocks, each with dimension $D = 64$ for output features. The number of clustered communities ($k_r$) was set to 7, inspired by the seven functional networks defined by Yeo et al. \cite{ThomasYeo2011}.

Our approach employed a 5-fold stratified cross-validation procedure on the ABCD dataset. In each fold, we divided the dataset into training, validation, and test sets, with the validation and test sets comprising 20\% of the original data. The neural network training was performed for 100 epochs, utilizing the Adam optimizer \cite{Adam} and the SmoothL1Loss loss function. An early stopping mechanism was incorporated, terminating training if the validation loss failed to decrease over 30 consecutive epochs. Additionally, the learning rate was reduced by a factor of 0.3 with a patience setting of 30. We utilized a batch size of 32.

We compared our proposed model with some baseline models, which included BrainGNN \cite{Li2021}, BrainNetCNN \cite{BrainNetCNN}, BrainRGIN \cite{Thapaliya2025},  Graph Transformer (GT) \cite{GraphTransformer}, and Support Vector Regression (SVR).

\subsection{Results}

In this section, we present the results of evaluating the performance of our proposed (\technique) against several baseline models on two key tasks: predicting total cognitive scores and classifying gender based on static FC data from the ABCD dataset. The comparison focused on demonstrating the superiority of our novel attention mechanism over the traditional Graph Transformer (GT) model by highlighting improved performance metrics.

\subsubsection{Total Cognitive Score Prediction}

Table \ref{tab:baseline-comparison} presents the comparative results for the total cognitive score prediction task. Our model, \technique, achieved an average Mean Squared Error (MSE) of 271.02 and an average correlation coefficient of 0.283. These results not only surpass the performance of the vanilla Graph Transformer but also outperform other state-of-the-art models such as BrainGNN \cite{Li2021} and BrainNetCNN \cite{BrainNetCNN}.

\begin{table}[ht]
\caption{Comparison of \technique\ with baseline models for total cognitive score prediction}
\label{tab:baseline-comparison}
\resizebox{\columnwidth}{!}{%
\begin{tabular}{|l|c|c|c|}
\hline
\textbf{Models} & \textbf{Avg. MSE} & \textbf{MSE $\pm$ Std. Dev.} & \textbf{Avg. Correlation} \\ \hline
\technique\ (Ours) & \textbf{271.02} & $271.02 \pm 3.21$ & \textbf{0.283} \\ \hline
Graph Transformer (GT) \cite{GraphTransformer} & 283.43 & $283.43 \pm 2.55$ & 0.25 \\ \hline
BrainGNN \cite{Li2021} & 272.02 & $273.02 \pm 3.01$ & 0.28 \\ \hline
BrainNetCNN \cite{BrainNetCNN} & 278.02 & $278.02 \pm 2.98$ & 0.25 \\ \hline
Support Vector Regression (SVR) & 280.12 & $280.12 \pm 2.34$ & 0.25 \\ \hline
\end{tabular}%
}
\end{table}

\technique\ demonstrates a significant improvement over the vanilla GT, achieving a lower MSE by approximately 12.41 units and a higher correlation coefficient by 0.03. This improvement underscores the effectiveness of our novel attention mechanism, which is specifically designed to handle subnetwork structures within brain graphs. By introducing a self-clustering approach, \technique\ allows for specialized attention mechanisms tailored to distinct functional clusters, thereby enhancing the model's ability to capture intricate brain connectivity patterns.

\subsubsection{Gender Classification}

To further validate the effectiveness of our proposed attention mechanism, we conducted a gender classification task comparing \technique\ with the vanilla GT. The results are summarized in Table \ref{tab:gender-classification}.

\begin{table}[ht]
\caption{Comparison of \technique\ with vanilla Graph Transformer for gender classification}
\label{tab:gender-classification}
\centering
\begin{tabular}{|l|c|c|}
\hline
\textbf{Models} & \textbf{Accuracy (\%)} & \textbf{F1-Score} \\ \hline
\technique\ (Ours) & \textbf{84.5} & \textbf{0.86} \\ \hline
Graph Transformer (GT) \cite{GraphTransformer} & 82.3 & 0.84 \\ \hline
\end{tabular}
\end{table}

In the gender classification task, \technique\ achieved an accuracy of 84.5\% and an F1-score of 0.86, outperforming the vanilla Graph Transformer, which attained 82.3\% accuracy and an F1-score of 0.84. This marginal yet significant improvement further validates the advantage of our self-clustering attention mechanism in enhancing the model's discriminative capabilities.

\section{Interpretation of Learned Subclusters}
\label{sec:interpretation}

A notable advantage of our \technique\ is its ability to learn and interpret subclusters within the brain's functional connectivity network. This interpretability is facilitated by the shared parameter matrix $\theta^{(l)}_{1}$, which captures the association between each of the 100 brain regions (nodes) and the 7 predefined functional communities. To identify clusters from the $(100 \times 7)$ $\theta^{(l)}_{1}$ matrix, we assign each node to the community with the highest association score in its corresponding row. Mathematically, for each node $i$, the assigned cluster $c_i$ is given by:

\[
c_i = \underset{c \in \{1,\dots,7\}}{\arg\max} \, \theta^{(l)}_{1}[i, c]
\]

This approach ensures that each brain region is uniquely assigned to the most strongly associated functional community, facilitating clear and interpretable clustering of the brain's functional networks.

\begin{figure}[ht]
    \centering
    \includegraphics[width=0.8\columnwidth, height=6cm]{ 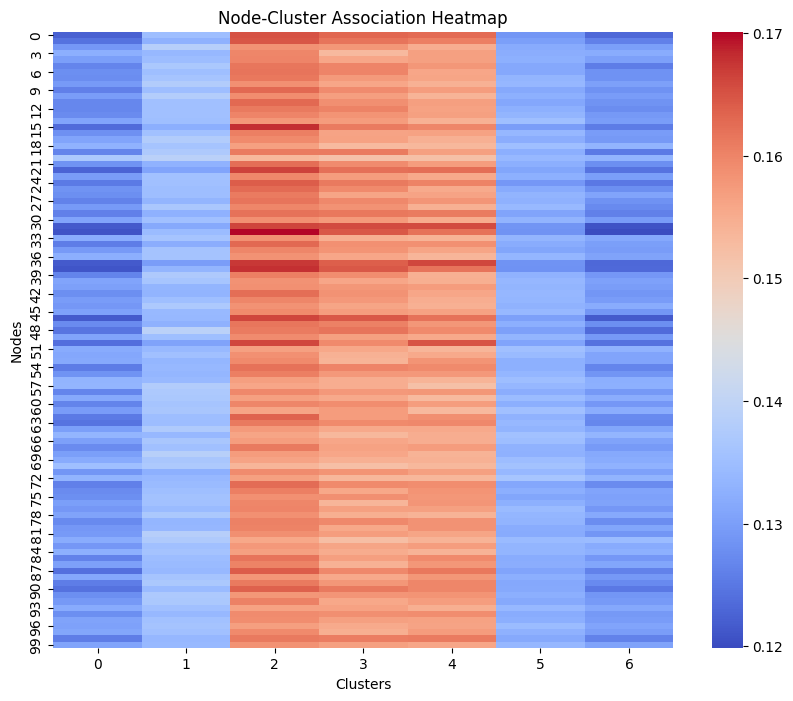}
    \caption{Visualization of the learned $\theta^{(l)}_{1}$ matrix. Each row represents a brain region, and each column represents a functional community. Higher values indicate stronger associations.}
    \label{fig:theta1_matrix}
\end{figure}

The intensity of each cell indicates the strength of association between a node and a community. By assigning each node to the community with the highest score, we obtain distinct functional clusters that align with known brain networks, such as the default mode network or frontoparietal network \cite{ThomasYeo2011}.

\section{Discussion and Conclusion}

In this research study, we developed a novel technique called \technique, which enhances the attention mechanism of the transformer by using self-clustering graph attention which enables nodes belonging to distinct clusters representing different brain networks to learn embeddings in a customized manner, addressing the issue of uniform node updates in traditional GNNs. Our preliminary experiments on the ABCD dataset demonstrate that \technique\ outperforms existing models in predicting total cognitive scores and gender classification, indicating its potential for modeling brain functional connectivity and interpreting underlying subnetwork structures.

This is preliminary work, and in the future, we aim to perform more extensive experiments on other phenotypes and explore the interpretability of the learned subclusters from \technique. Furthermore, this approach can be extended to make it usable for any kind of graphs with subnetwork, and we plan to test on general graph benchmark datasets to see if subclustered updates are useful in overall scenarios.

\section{Acknowledgments}
\label{sec:acknowledgments}

This study was funded in part by NIH grant R01AG063153.

\bibliographystyle{IEEEbib}
\bibliography{ refs}

\end{document}